\pdfoutput=1
\documentclass[11pt,a4paper]{article}
\usepackage[hyperref]{emnlp2018}
\usepackage{times}
\usepackage{latexsym}
\usepackage{url}
\usepackage{xspace}
\usepackage{graphicx}
\usepackage{amsmath,amssymb}
\usepackage{bm}
\usepackage{booktabs}
\usepackage{background}
\usepackage[most]{tcolorbox}

\title{Linguistic Typology Features from Text: Inferring the Sparse Features of World Atlas of Language Structures}

\author{Alexander Gutkin \enspace\enspace Tatiana Merkulova \enspace\enspace Martin Jansche\\
  Google Research, London, United Kingdom\\
  {\tt \{agutkin,merkulova,mjansche\}@google.com}}

\date{}

\aclfinalcopy 

\newcommand*{\eg}{e.g.\@\xspace}
\newcommand*{\ie}{i.e.\@\xspace}

\tcbset{textmarker/.style={%
        enhanced,
        parbox=false,boxrule=0mm,boxsep=0mm,arc=0mm,
        outer arc=0mm,left=6mm,right=3mm,top=7pt,bottom=7pt,
        toptitle=1mm,bottomtitle=1mm,oversize}}
\newtcolorbox{noteblock}{textmarker,
    borderline west={6pt}{0pt}{blue},
    colback=blue!10!white}

\definecolor{textcolor}{HTML}{778899}
\newcommand\Text{Preprint not submitted to EMNLP 2018 (prepared: April, 2018).}
\SetBgColor{textcolor}
\SetBgOpacity{1}
\SetBgAngle{90}
\SetBgPosition{current page.center}
\SetBgVshift{-0.54\textwidth}
\SetBgScale{1.1}
\SetBgContents{\sffamily\Text}

\begin{document}
\maketitle

\begin{abstract}
The use of linguistic typological resources in natural language processing has
been steadily gaining more popularity. It has been observed that the use of
typological information, often combined with distributed language representations,
leads to significantly more powerful models. While linguistic typology
representations from various resources have mostly been used for conditioning
the models, there has been relatively little attention on predicting features
from these resources from the input data. In this paper we investigate whether
the various linguistic features from World Atlas of Language Structures (WALS)
can be reliably inferred from multi-lingual text. Such a predictor can be used
to infer structural features for a language never observed in training data.
We frame this task as a multi-label classification involving predicting the
set of non-mutually exclusive and extremely sparse multi-valued labels
(WALS features). We construct a recurrent neural network predictor based on
byte embeddings and convolutional layers and test its performance on 556
languages, providing analysis for various linguistic types, macro-areas,
language families and individual features. We show that some features from
various linguistic types can be predicted reliably.
\end{abstract}

\begin{noteblock}\small
\emph{Note from the authors (April, 2020):} The goal of this work was to
investigate how far we can get from the character signal alone,
without recourse to the more informative mechanisms, such as
conditioning on word embeddings. The analysis of the results raised
several questions regarding the design of the experiment: using the
character input features alone how can one possibly predict phenomena
like double-headed relative clauses or the morphological signaling of
negation (Table~\ref{tab:metrics_by_feature}) with such reliability?
Is the system learning or just memorizing? A further possible
confounding factor is the extreme sparsity of WALS features, which may
not be modeled adequately by the weighted cross-entropy loss
(Section~\ref{sec:models}) leading to possible classification
randomness. We believe a significantly more focused and controlled set
of experiments is required to address these concerns.
\end{noteblock}

\section{Introduction}
The field of linguistic typology organizes the world's languages according to their
structural and functional features and helps to describe and explain the linguistic
diversity~\cite{song2013}. In recent years there has been a growing interest in employing
linguistic typology resources in natural language processing~\cite{asgari2017},
where one of its primary applications has been work towards scaling up the existing
language technologies to the long tail of world's languages~\cite{ohoran2016} for
which the traditional resources are very scarce or missing altogether.

Typological resources such as PHOIBLE~\cite{phoible}, Glottolog~\cite{glottolog2018},
PanPhon~\cite{mortensen2016} and World Atlas of Language Structures
(WALS)~\cite{wals2013} have been successfully used in diverse speech and language
tasks such as grapheme-to-phoneme conversion~\cite{peters2017}, multilingual
language modeling~\cite{tsvetkov2016polyglot}, dependency parsing~\cite{ammar2016}
and text-to-speech~\cite{tsvetkov2016linguistic}.

In this work we investigate the task of learning linguistic typological information
from multilingual text corpora. We frame this problem as a text classification
task where, given a certain text of arbitrary length, one needs to determine
the structural features of the corresponding language. The source for the features
is the World Atlas of Language Structures (WALS)~\cite{wals2013} that contains
phonological, lexical, grammatical and other attributes gathered from descriptive
materials for 2,679 languages.

Impressive progress has been achieved in the field of language
identification~\cite{lui2014} and accurate models are available for a large number
of languages~\cite{jauhiainen2017}. For the task at hand, however, in certain
situations one cannot simply look up the required features by the language code
having run the text through a language identifier. Such situations arise when
insufficient amounts of training data are available for a language or no training
data is available at all. As a hypothetical example, when applied to a text in
Scots or Cornish, the language identification will assign the text to English and
Welsh, respectively. This is not very helpful because, for this task, one is
interested in the linguistic features that make Scots unique -- possibly, as a
stretch, it's phonological relation to Old Norse~\cite{heddle2010},
or, in the case of Doric Scots, the Norwegian influence from the time of
late Middle Ages~\cite{lorvik2003}.

This task is attractive because having an accurate linguistic typology detector
can aid development work in speech and language fields. Correctly identifying
broad phonetic features of an unknown language can help one build crude
grapheme-to-phoneme rules and phoneme inventories for automatic speech recognition
and text-to-speech. Simple morphological and syntactic analyzers can potentially be
constructed given the knowledge of core syntactic and morphological
attributes, such as subject-verb agreement, word order or gender categories.
Given the hypothesis that basic word order and prosody are
correlated~\cite{bernard2012}, if the basic word order features can be reliably
inferred, one can construct prosodic models of prominence. Finally, such models
can potentially aid measuring the linguistic change. For example, given a text in
17th century Romani~\cite{matras1995} (and assuming that it gets overall
classified as modern Romani) one can possibly observe the features that it
acquired or lost in the course of three hundred years.

The focus on this study is whether a reliable WALS feature neural network classifier
can be constructed from multilingual, not necessarily parallel, text
corpora. We describe the related typology prediction work
in Section~\ref{sec:related} and introduce our approach to predicting WALS
features as a multi-label classification problem~\cite{gibaja2014}. It's worth
noting that this task is different from the related work reported in the
literature~\cite{malaviya2017,bjerva2018}. First, the goal of the classifier
is to correctly infer the sparsely defined WALS features, rather than filling
the missing gaps. Second, the classifier is trained on open text, rather than
language embeddings produced as by-product of some other task (\eg, from
parallel machine translation corpora). Finally, we make no assumptions about
the input language of the text and employ no language identifying input features.

The goal of the experiments, described in Section~\ref{sec:experiments}, is to
determine which features and groups thereof can be reliably inferred. In addition,
we provide some of the results for various languages and their phylogenetic
groupings. Since the task is reasonably novel, our goal is to provide a baseline,
a very likely crude one, but one that can be gradually improved upon over time.
%
\section{Related Work and Preliminaries}
\label{sec:related}
\paragraph{Related Work:}
A recent popular approach is to represent languages as dense real-valued vectors,
referred to as \emph{language embeddings}. It is assumed that these distributed
language representations implicitly encode linguistic typology information.
The language embeddings can be obtained by training a recurrent neural
network language model~\cite{mikolov2010} jointly for multiple
languages~\cite{tsvetkov2016polyglot,ostling2017}. Alternatively, the embeddings
can be trained as part of other tasks, such as part-of-speech tagging~\cite{bjerva2018}
or neural machine translation~\cite{malaviya2017}.

\newcite{malaviya2017} note that existing typological databases, such as WALS,
provide full feature specifications for only a handful of languages. In order to
fill this gap they construct a massive many-to-one neural machine translation
(NMT) system from 1017 languages into English (relying on a parallel database of
biblical texts) and use the resulting language embeddings to succesfully
predict the missing information for the under-represented languages.

\newcite{bjerva2018} produce language embeddings in the process of training a
part-of-speech tagger for Uralic languages. At various stages of the training
the authors constructed a logistic regression model that takes language embedding
as an input and outputs a typological class the language belongs to according to
WALS. They found that certain WALS features could be inferred from the embeddings
with accuracy well above the baseline.
\paragraph{Multi-Label Classification:}
Given an example text representation in an input feature space, $\mathbf{x} \in X$,
the classification task consists of selecting a set of multiple applicable WALS
feature labels $\{ \lambda_i \}$ from a finite set of labels
$L = \{ \lambda_1, \lambda_2, \ldots, \lambda_{N_L} \}$, where $N_L$ is the number of WALS
features (192 for the WALS version used in this work). Each candidate
label takes its value from a set of disjoint classes $Y_i = \{ y^i_j \}, 1 \leq i \leq N_L$,
corresponding to the values of a particular WALS feature $\lambda_i$.
For example, language may or may not have a \textsc{number of genders} feature label
present, but if it is present this feature cannot take the values of \textsc{None} and
\textsc{Four} simultaneously. This scenario fits the \emph{multi-label multi-class}
classification problem~\cite{gibaja2014,zhang2014,madjarov2012,yang2009}. Examples
of natural language processing tasks where this type of problems arise
is sentiment analysis~\cite{liu2015} and text classification~\cite{pestian2007}.
\paragraph{Data Imbalance:}
As noted by~\newcite{malaviya2017} and~\newcite{littell2017}, many typological
databases are designed to suit the needs of theoretical linguistic typology, resulting
in a sparse representation of features across languages (mostly due to intentional statistical
balancing of features across language families and geographic areas). Also, for certain
languages the maintainers are sometimes unable to obtain reliable description of linguistic
attributes from the available linguistic sources, \eg~\newcite{comrie2009}. This situation is
not perfect for statistical modeling because it results in heavy \emph{data imbalance}
between different types of features and complicates construction of machine
learning models. A classifier constructed without regard to data imbalance will lean towards
correctly predicting the majority class, which in case of WALS corresponds to missing
or intentionally undefined features, while the ``interesting'' features with low coverage will
receive corresponding proportion of classifier's attention.

Approaches to data imbalance have been extensively studied in the literature~\cite{japkowicz2002,he2009,krawczyk2016}
and the proposed remedies include altering the training data balance by \emph{upsampling}
(replicating cases from the minority), \emph{downsampling} (removing cases from the majority),
synthetically generating cases~\cite{he2009,garcia2016} and design of special
metrics~\cite{charte2015}. In this study we choose to keep the original training data as is
in order not to disturb the original gentle balance between language representations and
instead adjust the classifier optimization algorithm~\cite{king2001} as well as introducing
special logic for decoding the logits into posterior probability estimates.
%
\section{Corpora}
\label{sec:corpora}
\paragraph{Text Corpus:} For the training text data we used the second
release of LTI LangID language identification corpus from
CMU.\footnote{\scriptsize\url{http://www.cs.cmu.edu/~ralf/langid.html}}
The core corpus contains training data for 847 languages, and some (possibly very tiny)
amount of text for a total of 1146 languages. The data predominantly comes from
Wikipedia text and many of the Bible translations (redistributable under Creative Commons
licenses) as well as Europarl corpus of european parliamentary proceedings~\cite{koehn2005}.
The core subset of the corpus contains languages for which sufficient text
is available to generate quality models. \newcite{brown2014non} notes that there
is no fixed minimum amount to be included in this category; generally, Bibles
require less text than Wikipedia and languages with few lexically-similar languages
require less than those with much lexical overlap. In this study we focus on the
core subset of the corpus and treat various dialects of the same language (if
these are present in the corpus) as distinct languages. The corpus is divided
into training, development and test subsets, with the individual examples ranging
from short words to single sentences and whole paragraphs.
%
\begin{figure}
\centering
\includegraphics[width=0.32\textwidth]{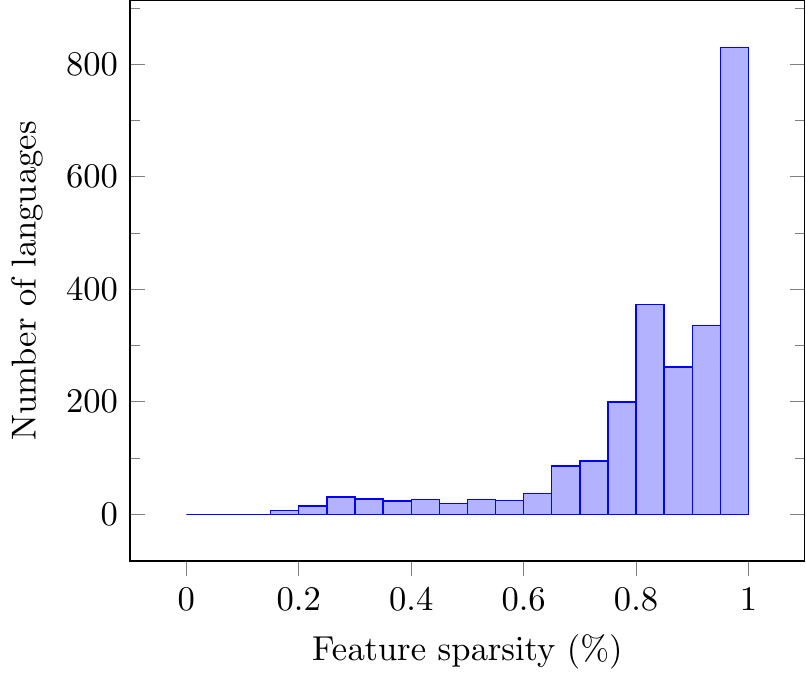}
\caption{WALS features sparsity estimated as percentage of the features
not attested for each language.}
\label{fig:wals_sparsity}
\end{figure}
\paragraph{Typology Dataset:} We use World Atlas of Language Structures
(WALS)~\cite{wals2013} as a source of typological information for 2,679
languages. The 192 WALS typological multi-valued language features are
organized into 152 chapters, each chapter corresponding to a particular
phonological, morphological or syntactic linguistic property. For example,
there are 18 chapters corresponding to \textsc{Word Order} syntactic property
where most chapters contains one typological feature, such as
\textsc{Order of Genitive and Noun}, while other chapters, such as
\textsc{Order of Negative Morpheme and Verb} contain 7 features, such as
\textsc{Obligatory Double Negation}~\cite{dryer2013neg}. The dataset is very
sparse (Figure~\ref{fig:wals_sparsity}). For example, for 1,801 languages
out of 2,679 only 20\% (or less) of the WALS features are attested.
Only 149 languages have 50\% (or more) coverage.
\paragraph{Preparing the Data:} First, we prune WALS by removing the languages
for which no ISO 639-3 code is defined. This set includes 57 languages. In addition,
we remove three languages for which no WALS features are attested. The training,
development and test sets for neural network classifier are constructed by
assigning to each example in LTI LangID dataset the corresponding WALS features.
We match each LTI LangID example with the WALS features using the ISO 639-3
language code. Some LTI LangID language codes are in two-letter ISO 639-1
format which we convert to ISO 639-3 before attempting the lookup. Using this
procedure we drop 338 languages for which no corresponding WALS entry can be
reliably located using the ISO language code. The resulting dataset consists of 544
languages in the training set, 108 languages in the development set and 556
languages used for testing.
%
\section{Methodology and Models}
\label{sec:models}
The goal of this study is to construct a model for inferring WALS features from
text in many languages. The classifier design makes no special assumptions about
code switching in the input text and does not use language identifying features
rather than the text itself.
%
%
\subsection{Embedding Layer}
After combining the CMU LTI LangID and WALS datasets (described in
Section~\ref{sec:corpora}) one is left with about 6,2M input sequences in
544 languages. Tokenizing the training sequences into words and training
parametrization of words as vectors, known as word embeddings~\cite{mikolov2010},
to be used as inputs to the neural network classifier is not going to work
well because the amount of data at hand is not sufficient. In addition,
some languages, such as Khmer and Burmese, require segmentation to obtain
words, which in itself is a hard problem~\cite{ding2016word}. Furthermore,
the word embeddings don't provide us with a flexible way of capturing similar
words in morphologically rich languages. To make this problem more tractable
we employ character-level embeddings, which require fewer parameters than
word-level embeddings and need no special preprocessing, such as complex
tokenization~\cite{zhang2015,kim2016,wietling2016,irie2017}. Similar to one of the
competitive representations reported by~\newcite{zhang2017} and unlike other
approaches that operate on Unicode code points, \eg~\cite{jaech2016}, we
decompose the input text into UTF-8 byte sequences, which include white
space characters. We consider two ways to model the UTF-8 byte-level embeddings.

\paragraph{Byte Unigrams:} Let $\mathbf{x} = (x_1, x_2, \ldots, x_T)$ be a byte
representation of an input sequence. In the simplest scenario, similar to~\newcite{xiao2016},
we treat each byte as a separate unigram from a small vocabulary $V$ consisting of
256 values with the addition of an end-of-sentence and padding markers. At time
$t$, byte input $x_t$ is one-hot encoded into a vector $\mathbf{c}_t$ and
multiplied with the embedding matrix $W_c \in \mathbb{R}^{\vert V \vert \times d}$
to produce a $d$-dimensional embedding vector $\mathbf{e}_t$.

\paragraph{Byte $n$-grams:} We also investigate the byte $n$-gram embeddings,
where instead of individual bytes, the text is transformed into a sequence of
UTF-8 byte $n$-grams. In other words,
$\mathbf{x} = (\mathbf{x}^1_1, \mathbf{x}^2_1, \ldots, \mathbf{x}^n_1, \ldots,
\mathbf{x}^{T-1}_{T-1}, \mathbf{x}^T_{T-1}, \mathbf{x}^T_T)$, where $\mathbf{x}_i^j$
denotes a subsequence of bytes in $\mathbf{x}$ from position $i$ to position $j$
inclusive, \ie $\mathbf{x}_i^j = (x_i, x_{i+1}, \ldots, x_j)$, where
$\mathbf{x}_i^i = x_i$ and $n$ is the maximum length of an $n$-gram window,
decomposition similar to~\cite{wietling2016}. In this approach, the decomposition
has the effect of lengthening the original byte sequence. We compute the
$d$-dimensional embedding $\mathbf{e}_t$ for an $n$-gram $\mathbf{x}^{t+k}_t$,
$1 \leq k \leq n$, at time $t$ as
$\mathbf{e}_t = \frac{1}{k}\sum_{i=1}^k \mathbf{c_i} W_c$, where $\mathbf{c}_i$
is a one-hot encoding of byte $x_i$ and $W_c$ is the embedding matrix. Because
$n$-grams are represented by individual byte aggregation, the dimensions of the
embedding matrix can be compact, similar to the byte unigram representation.

\paragraph{Character $n$-grams as words:} In this approach, the input text is
transformed into sequences of Unicode character, rather than byte, $n$-grams.
Each $n$-gram is treated as a unique undecomposable word from a possibly large
vocabulary $V$ and embeddings are constructed similarly to word embedding
approaches~\cite{mikolov2013}.
\subsection{Convolutional and Recurrent Layers}
Given the embeddings, they can be treated as a kind of raw signal at character level to
which one can apply one-dimensional temporal convolutions to extract important local
context features. First introduced by~\newcite{zhang2015}, this approach has proven
to be competitive to models built on word embeddings~\cite{kim2016,irie2017}. We are
adopting the same multiple convolutional layer configuration as the one reported
by~\newcite{xiao2016}. Applying the dropout~\cite{srivastava2014} to the outputs
of the embedding layer as well as the final convolution layer turned out to be
effective.

Similar to others~\cite{kim2016,xiao2016,josefowicz2016,vosoughi2016}, we experiment
with a hybrid architecture, where the outputs of a convolutional neural network (CNN)
are used as inputs to a recurrent neural network (RNN). In our experiments, for an RNN we
employ a bidirectional variant~\cite{graves2005} of long short term memory (LSTM)
model~\cite{hochreiter1997}, with an application of dropout.
\subsection{Logits Layer and Optimization Strategies}
We have looked into two approaches to optimization. In the first approach, we treat
the problem as a standard multinomial logistic regression, where at each time step
the network may output multiple non-exclusive labels. The forward and backward
outputs corresponding to the last time step of a bidirectional LSTM are
concatenated together and fed into the single fully-connected linear activation layer.
Each output of this layer corresponds to a particular value of a WALS
feature. There are 1316 outputs in total. We apply sigmoid
non-linearities to the outputs of fully-connected layer and optimize all
predictions $\hat{\mathbf{y}}$ against the true labels $\mathbf{y}$ all at once
using cross-entropy function~\cite{bishop2006}
\begin{equation*}
L(\bm{\theta}) = - \frac{1}{N} \sum_{n=1}^N \sum_{i=1}^C
  \mathbf{y}^i_n \log
    \left(
      \hat{\mathbf{y}}^i(\mathbf{x}_n, \bm{\theta})
    \right) + r(\bm{\theta}) \ ,
\end{equation*}
where $\bm{\theta}$ represents network parameters, $\mathbf{x}$ are the training
sequences, $C$ is the dimension of the prediction vector and $r$ is an $l_2$-norm
regularization term. Recall that the task at hand is multi-label multi-class
classification. Since WALS features are non-exclusive
but their values corresponding to our predictions are not, we can only hope that
the network learns that within each feature the values are independent.

To address this potential shortcoming we also tested an alternative strategy which
constrains the universe of predicted values for each individual WALS feature to be
mutually independent. In this scenario, we break the problem down into 192
tasks, one for each WALS feature, where each individual problem is treated
as multinomial mutually-exclusive classification, somewhat similar to multi-task
learning~\cite{liu2016}. For each task, a fully-connected layer is constructed
that takes its input from the last time steps of the RNN and a softmax
non-linearity is applied to each layer. The loss function in this case is the
sum of individual softmax cross-entropy loss functions $L(\bm{\theta}_i)$ for
each task, $1 \leq i \leq 192$.
\subsection{Dealing with Data Imbalance}
\begin{figure}
\centering
\includegraphics[width=0.32\textwidth]{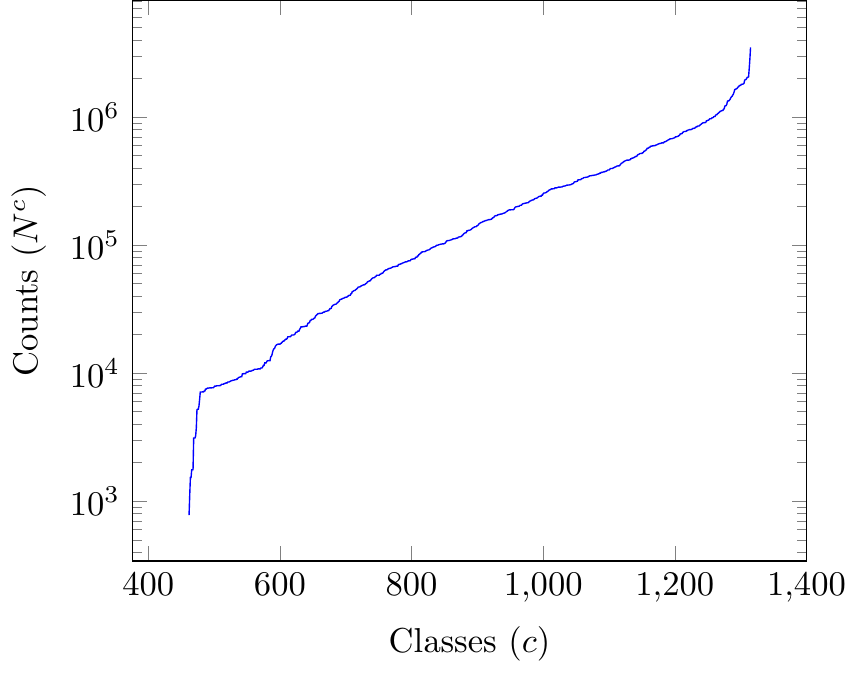}
\caption{WALS feature value (class) counts displayed on a logarithmic scale.
Classes are sorted by their counts. 461 classes out of 1316 are unobserved.}
\label{fig:dataset_class_counts}
\end{figure}
In Section~\ref{sec:corpora} we provided initial analysis of WALS feature
sparsity based on feature value counts computed solely from WALS corpus
(Figure~\ref{fig:wals_sparsity}). The dimension of difficulty involved in
training a neural network WALS feature classifier on CMU LTI LangID data
is demonstrated in Figure~\ref{fig:dataset_class_counts}, which shows the counts
for all possible WALS feature values (corresponding to classes that the classifier
has to predict) encountered in the training data. Significant proportion of
classes (461 out of 1316) is not encountered in the training data for 544
languages. The distribution of counts for the majority of the remaining classes
(approximately 700 in number) is approximately log-linear, while the remaining
155 classes are either very rare or very frequent.

To deal with this heavy class imbalance we employ the family of weighted
cross-entropy loss functions defined as
\begin{equation*}
L(\bm{\theta}) = - \frac{1}{N} \sum_{n=1}^N \sum_{i=1}^C
  w(i) \mathbf{y}^i_n \log
    \left(
      \hat{\mathbf{y}}^i(\mathbf{x}_n, \bm{\theta})
    \right) + r(\bm{\theta}) \ ,
\end{equation*}
where $w(c)$ is the weight function associated with class $c$ which is defined
as
\begin{equation*}
w(c) = \begin{cases}
         N_\mathcal{L} / \left(N_c M_\mathcal{L}\right) & \text{if $N_c>0$},\\
         0 & \text{if $N_c=0$}.
       \end{cases}
\end{equation*}
where $N_\mathcal{L}$ denotes the count of a WALS feature
$\mathcal{L}$ in the training data, $N_c$ is the feature value count
($c \in \mathcal{L}$) and $M_\mathcal{L} = \lvert \mathcal{L} \rvert$ is the
number of values for feature $\mathcal{L}$. This reciprocal frequency definition
of a weight function $w(c)$ above is inspired by~\newcite{king2001}. The purpose
of the function is to penalize the frequent classes and boost the rare ones. The
unattested classes do not contribute to the overall loss.\footnote{For the
multi-task approach we also tried to introduce label weights defined as
$N / (192 N_\mathcal{L})$ that scale the individual task loss functions,
but this modification did not lead to improvement in the models.}

Since 461 classes are not observed in the training data, additional modification
to the training regime consists of masking out the logits corresponding to these
classes before applying sigmoid or softmax (in the case of multi-task
optimization) non-linearities.
%
\section{Experiments}
\label{sec:experiments}
\begin{table*}[t]
\centering
\small%
\begin{tabular}{lcc|cccc|ccc}
\toprule
Type & $N_L$ & $N$ & \multicolumn{4}{c|}{Bytes ($\mathcal{B}$)} & \multicolumn{3}{c}{Characters ($\mathcal{C}$)} \\
     &       &     & $S^\mathcal{B}_{\max}$ & $N^\mathcal{B}$ & $\mu^\mathcal{B}$ & $\sigma^\mathcal{B}$ & $N^\mathcal{C}$ & $\mu^\mathcal{C}$ & $\sigma^\mathcal{C}$ \\
\midrule
\texttt{train} & 544 & 6,199,201 & 20.5K & 1,057M & 176.0 & 130.0 & 887M  & 147.5 & 108.5 \\
\texttt{dev}   & 108 & 79,856    & 9.9K  & 15.9M  & 199.4 & 152.7 & 12M   & 150.2 & 115.2 \\
\texttt{test}  & 556 & 226,235   & 7.2K  & 39.2M  & 174.7 & 131.4 & 32.8M & 146.0 & 109.7 \\
\bottomrule
\end{tabular}
\normalsize%
\caption{Dataset details showing, for each dataset type, the number of languages ($N_L$), the
total number of predictions ($N$) and some statistics computed on bytes ($\mathcal{B}$) and
characters ($\mathcal{C}$), respectively.}
\label{tab:dataset_details}
\end{table*}
\begin{table}
\centering
\includegraphics[width=0.48\textwidth]{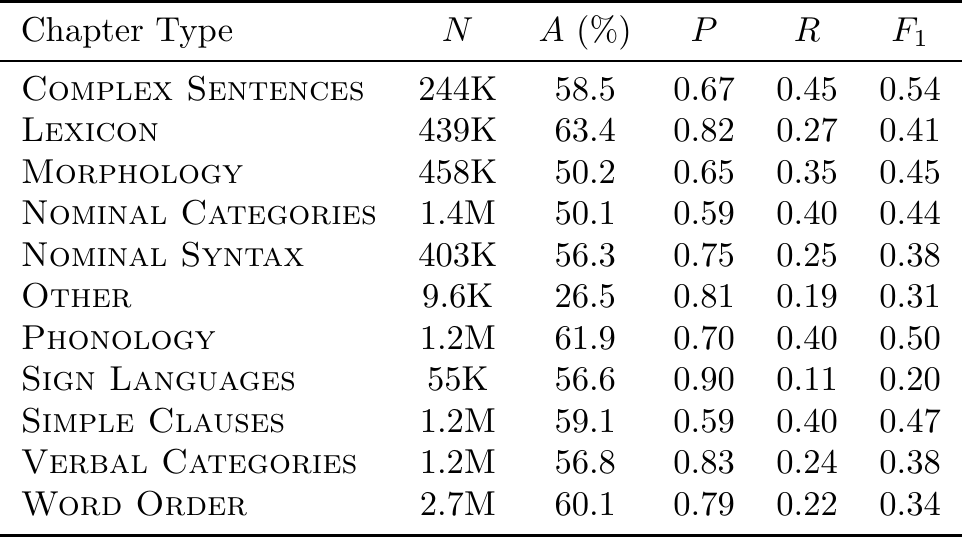}
\caption{Metrics for WALS features grouped by WALS chapter type.}
\label{tab:metrics_by_type}
\end{table}
\begin{table}
\centering
\includegraphics[width=0.48\textwidth]{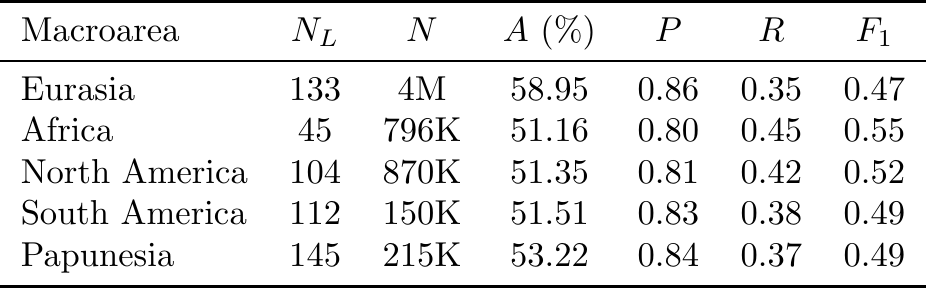}
\caption{Metrics for WALS features grouped by language macro-area.}
\label{tab:metrics_by_macroarea}
\end{table}
\begin{table*}
\centering
\includegraphics[width=0.95\textwidth]{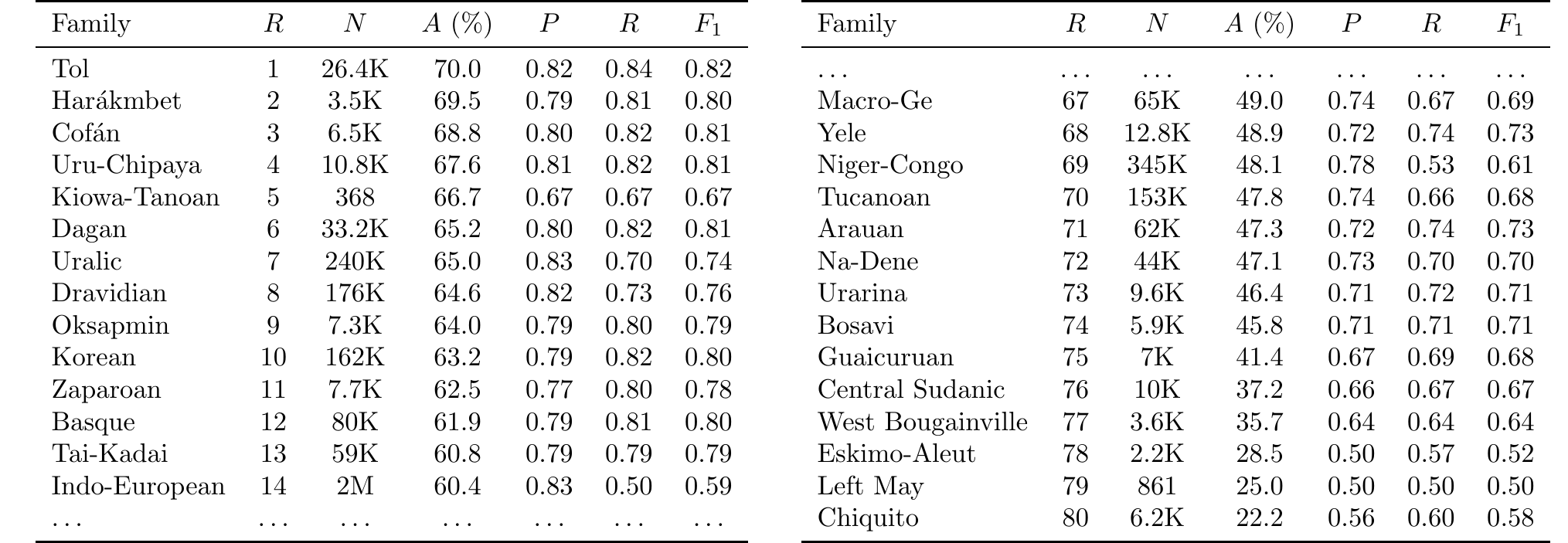}
\caption{WALS individual feature metrics grouped by 80 language families and ranked ($R$) by
accuracy ($A$). 14 best (left) and worst (right) scoring language families are shown.}
\label{tab:metrics_by_family}
\end{table*}
\begin{table}[t]
\centering
\includegraphics[width=0.49\textwidth]{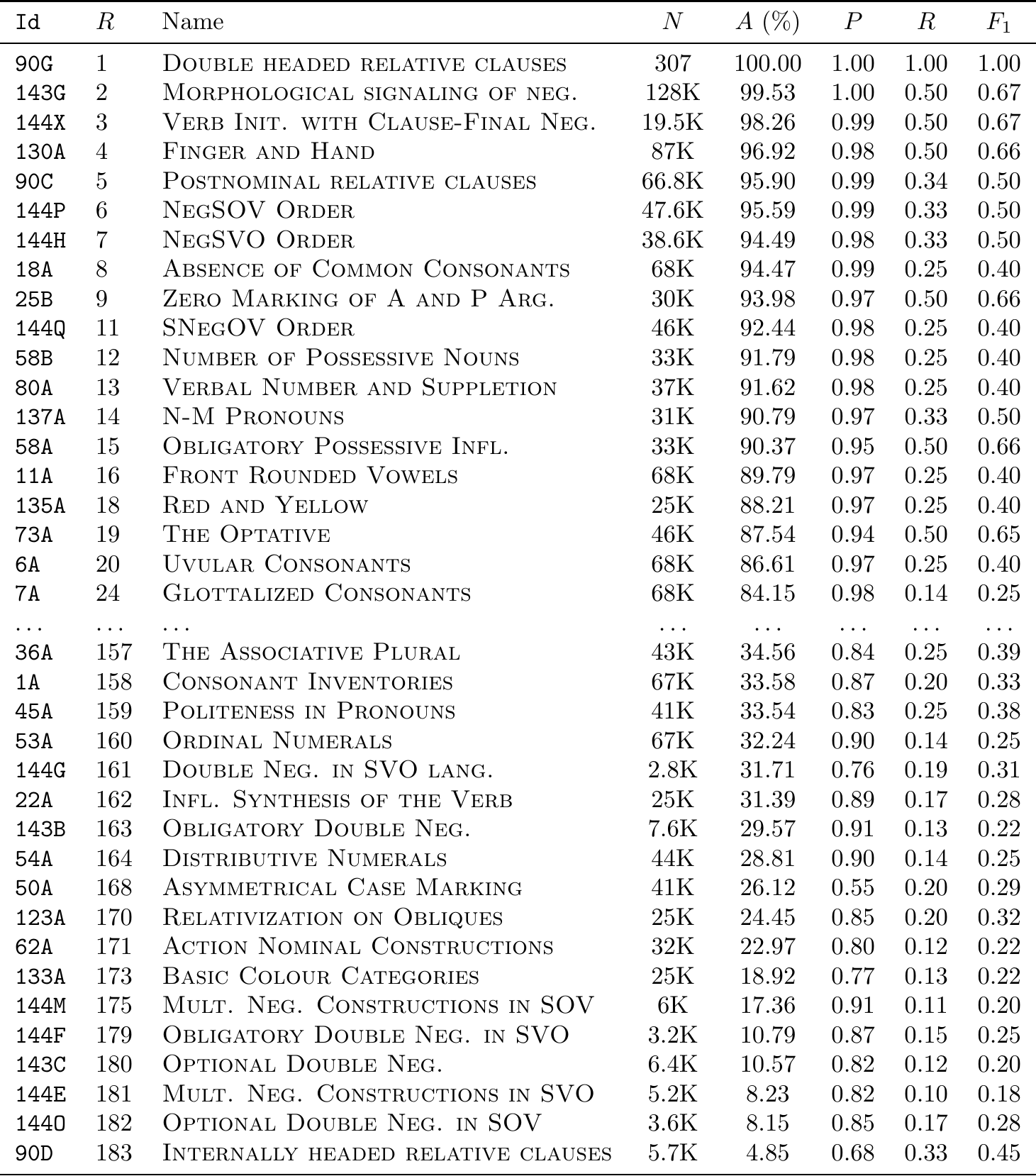}
\caption{Metrics for individual WALS features ranked ($R$) by accuracy ($A$) in descending order.}
\label{tab:metrics_by_feature}
\end{table}
\subsection{Dataset Preprocessing} The dataset details are shown in Table~\ref{tab:dataset_details}
where for training, development and test sets the number of languages and the corresponding
total number of sequences are shown. Statistics was computed on UTF-8 bytes ($\mathcal{B}$)
and Unicode characters ($\mathcal{C}$). One of the important indicators is the length of
individual sequences - some of the sequences correspond to short words while others represent
sentences or even the entire paragraphs. The presence of very long sequences is indicated by
the length (in bytes) of the longest sequence in each dataset (denoted $S^\mathcal{B}_{\max}$).
Instead of performing sentence splitting on individual sequences (which may be tricky
for languages with limited means of denoting sentence breaks) we retain all the
sequences which are between five and 600 characters long, omitting the rest from the
training and testing. The resulting total number of byte or character tokens and the mean
and standard deviation of sequence lengths are then computed for bytes
($N^\mathcal{B}$, $\mu^\mathcal{B}$, $\sigma^\mathcal{B}$) and characters
($N^\mathcal{C}$, $\mu^\mathcal{C}$, $\sigma^\mathcal{C}$). Pruning out the very
long sequences makes the training process more tractable by reducing the training
time, while retaining reasonably high variance in sequence lengths, as indicated by
the values of standard deviation.
%
\subsection{Network Architecture Details}
The experiments involve three network configurations, each corresponding to a
particular type of the embedding layer introduced in Section~\ref{sec:models}.
%
\paragraph{Embedding Layer:} The dimension $d$ of the
embedding vector is 8 for the individual byte embeddings, 32 for byte $n$-gram
embeddings and 256 for character $n$-gram embeddings. The dropout with probability
0.5 is applied to the embedding layer~\cite{srivastava2014}. The maximum length
of byte $n$-gram is 7, while for character $n$-grams we generate $n$-grams up to
the length of 5. The character $n$-grams are hashed in order convert the strings
into integer quantities. The number of hash buckets is set to $2^{14}$. When
training the embedding we initialize it using normal distribution
$(\mu_e,\sigma_e) = (0, \frac{1}{\sqrt{d}})$.
\paragraph{Convolutional Layer:}
The parameters for the convolutional are somewhat similar to one of configurations
in~\cite{xiao2016}: There are three one-dimensional convolution layers
containing 20, 40 and 60 filters, respectively. Receptive field sizes $r$ for
each layer are 5, 5 and 3. The stride parameter is set to 1. Rectified linear
units (ReLU) are used in each layer~\cite{glorot2011}. Batch normalization
is applied before each layer~\cite{ioffe2015}. Each convolution layer is followed by a
max-pooling layer with filter size $r'$ set to 2. Dropout with probability
0.5 is applied to the last max-pooling layer.
\paragraph{Bidirectional LSTM:} The RNN consists of stacked two-layer bidirectional
LSTM containing 128 cells each. Dropout is applied to each layer with probability
of 0.5. Uniform weight initialization scheme from~\cite{glorot2010} was used.
Residual connections are added to the second LSTM layer~\cite{wang2016:res}.
\paragraph{Optimization:} The models are trained using AdaDelta~\cite{zeiler2012}
with $\rho = 0.95$ and $\epsilon = 10^{-8}$. We use exponential learning rate
decay, with initial learning rate set to $5 * 10^{-5}$, reasonably slow
decay factor of $0.9$ and number of decay steps set to $3 * 10^5$. $L_2$
regularization is applied to the recurrent layer weights, with the weight
scaling factor set to $0.05$. In addition, value of 10 is used to clip
the global gradients. The training batch size is set to 8.
\subsection{Results and Analysis}
When computing the various metrics we ignore the undefined WALS feature values
focusing on attested features only, relying on the fact that during training
the weighted loss function alleviates the inherent imbalance between the WALS
classes. After pruning out the WALS features and the individual values
unattested in the training data, we are predicting 1316 possible values (classes)
of 183 WALS features (labels).\footnote{Due to space limitation, some of the
tables below contain partial results. The full tables are submitted
as supplementary material.}

\paragraph{Selecting the Best Model:} We used an accuracy metric computed on all
the WALS features encountered in the test set in order to select the best out
of the architectures described earlier. Our baseline was byte unigram LSTM-RNN
configurations with no convolutional layers for which an accuracy of 52.3\% was
achieved. We tested the configurations described in the previous section against
the baseline and found that the best performing architecture is a byte 7-gram
CNN-LSTM that achieves the accuracy of 57.2\% in the regular (non multi-task)
training mode. The character 5-gram configuration achieved a slightly worse
accuracy of 57.1\% and was also found to be more memory inefficient due to the
size of the embedding (which is necessary in order to treat character $n$-grams
as word-like units). A surprising discovery was that the multi-task-like training
did not perform as well as we had hoped with all the configurations scoring below
50\%. In addition, the multi-task training was significantly slower (taking one
day longer to converge) due to running numerically more complex optimization.

\paragraph{Chapter Types:}
Results for all the WALS features aggregated over chapter types are shown in
Table~\ref{tab:metrics_by_type}, where, for each chapter type, the total number of
predictions ($N$), accuracy ($A$), precision ($P$), recall ($R$) and $F_1$
scores are displayed. Despite reasonable precision values, the recall is
substantially lower for all the chapter types which is due to the high number of
predicted false negatives. The three chapter types with most accurate
(according to $A$) predictions are \textsc{Lexicon}, \textsc{Phonology} and
\textsc{Word Order}. These contain lexical, phonological and word
order-related features. Interesingly, the least accurate chapter type
is \textsc{Morphology}, even though we intuitively expect the results for
the morphological features to be on par with the lexical features.

\paragraph{Language Macro-Areas:}
Another informative comparison is to group the predictions over WALS linguistic
macro-areas~\cite{hammarstrom2014} shown in Table~\ref{tab:metrics_by_macroarea},
where $N_L$ is the number of languages tested for the particular macro-area,
$N$ is the total number of predictions, and metrics, similar to the ones
employed when aggregating over chapter types, are shown. The best results
according to accuracy ($A$) and precision ($P$) correspond to the languages
of Eurasia. We hypothesize that this can potentially be explained by two
factors: First, the proportion of the training data for Eurasian languages
is significantly higher than for the other languages and, second, the Eurasian
languages are likely to be better documented, resulting in more detailed
WALS descriptions and hence lower feature sparsity.

\paragraph{Language Families:}
Table~\ref{tab:metrics_by_family} shows the results aggregated over 28 language
families out of 80, where the 14 languages on the left correspond to the
best performing group and the 14 languages on the right to the worst performing
group (according to accuracy $A$). Language rank is $R$ and $N$ denotes the
total number of predictions. With the exception of a couple of outliers,
most of the top performing languages have relatively high and balanced values
of precision and recall. Interestingly enough, the top most accurate language
families correspond to very small languages of South and Central America. In
the case of Tol, Har{\'a}kmbet and Uru-Chipaya, the families consist of a single
language spoken by around a thousand (or less) speakers. Predictions for these
very low-resource languages are significantly more accurate than for some much larger
and better documented families in the list, such as Uralic, Dravidian, Tai-Kadai
and Indo-European, although the precision and recall values are overall roughly
in the same range. Among the poorly scoring (in terms of accuracy) language
families shown in the table on the right, the poor scores for Niger-Congo and
Central Sudanic language families can be singled out. The result for Niger-Congo
family is especially disappointing because this is one of the major language
families both in terms of number of distinct languages and the number of speakers.
Despite low accuracy, however, the precision value of 0.78 for Niger-Congo family
is reasonable.

\paragraph{Individual Features:}
Table~\ref{tab:metrics_by_feature} shows various metrics for the short (best) head
and long (worst) tail of 183 individual WALS features, ranked by accuracy ($A$).
For each feature, the corresponding WALS feature identifier (\texttt{Id}), its
rank ($R$), name and the number of predictions ($N$) is shown along with the corresponding
metrics. For most of the features, precision completely dominates the recall due
to high number of false negatives. The accurately predicted features (with accuracy
over 80\%) and poorly predicted ones come from diverse WALS chapter types with no
clear ``winning'' type to declare. For example, both \textsc{NegSOV Order} and
\textsc{NegSVO Order} (from \textsc{Word Order} chapter) are reliably and very
well predicted, while the prediction of \textsc{Optional Double Negation in SOV}
feature from the same chapter type is extremely poor. The same observation holds for
other chapter types, such as \textsc{Nominal Syntax}. It is interesting to note that
some features from the \textsc{Phonology} type, such as \textsc{Front Rounded Vowels},
are among the most accurate.
%
%
\section{Conclusion}
In this study we approached the problem of predicting the attested sparse WALS
features as a multi-label classification problem. We have shown that by building
a resonably standard recurrent neural network classifier following the recipes from
the existing literature, combined with a simple reciprocal frequency weighting for
alleviating the class imbalance, we can reliably predict at least some of individual
WALS features. An interesting finding that confirms the finding of~\newcite{malaviya2017}
is that the features come from a variety of linguistic types. Despite these
promising initial findings, much work still remains: We need more sophisticated
techniques, such as SMOTE~\cite{jeatrakul2010}, to make classifier more robust
against the WALS feature sparsity. Furthermore, to mirror some of the conclusions
of~\newcite{wietling2016}, in our situation a simpler architecture, perhaps not
even a neural one, may have performed better than state-of-the-art CNN-LSTM hybrid
model.\footnote{We will be releasing the code used in the experiments into public domain.}
In addition to improving our models, we would also like to investigate the
correlations between different groups of WALS features, provide a more in-depth
typological analysis for performance of various features and test our models against
the languages not seen in the training data.

\bibliography{text_wals}
\bibliographystyle{acl_natbib_nourl}

\end{document}